\documentclass[sigconf]{acmart}
\usepackage{multirow}
\usepackage[utf8]{inputenc} 
\usepackage[T1]{fontenc}    
\usepackage{url}            
\usepackage{booktabs}       
\usepackage{amsfonts}       
\usepackage{nicefrac}       
\usepackage{xcolor}         

\usepackage{mathtools}
\usepackage{amsthm}
\usepackage{colortbl}
\usepackage{tablefootnote}
\usepackage{booktabs}
\usepackage{adjustbox}
\usepackage{wrapfig}
\usepackage{graphicx}     
\usepackage{pifont}  
\usepackage{hhline}
\usepackage{enumitem}
\usepackage{etoolbox}
\usepackage{listings}
\makeatletter  
\newcommand{\thickhline}{%
    \noalign {\ifnum 0=`}\fi \hrule height 1pt
    \futurelet \reserved@a \@xhline
}
\definecolor{mygray}{gray}{.9}
\definecolor{ggray}{RGB}{127,127,127}
\definecolor{reda}{RGB}{192,0,0}
\definecolor{redb}{RGB}{217,148,143}
\definecolor{myyellow}{RGB}{190,144,0}
\definecolor{mygreen}{RGB}{0,176,80}
\definecolor{myred}{RGB}{248,66,0}
\definecolor{myblue}{RGB}{30,90,100}
\definecolor{mygray1}{RGB}{245,245,245}

\definecolor{agent}{RGB}{218, 165, 32}
\definecolor{goods}{RGB}{220, 20, 60}
\definecolor{payment}{RGB}{34,139,34}
\definecolor{seller}{RGB}{70,130,180}
\definecolor{place}{RGB}{138,43,226}

\usepackage{xspace}
\usepackage{booktabs}
\usepackage{multirow}
\usepackage[table]{xcolor}
\usepackage{caption}
\usepackage{pifont}
\usepackage{subcaption} 
\AtBeginDocument{%
  }

\copyrightyear{2026}
\acmYear{2026}
\setcopyright{cc}
\setcctype{by}
\acmConference[MM '26]{Proceedings of the 34th ACM International Conference on Multimedia}{November 10--14, 2026}{Rio de Janeiro, Brazil}
\acmBooktitle{Proceedings of the 34th ACM International Conference on Multimedia (MM '26), November 10--14, 2026, Rio de Janeiro, Brazil}
\acmDOI{10.1145/3767308.3836490}
\acmISBN{979-8-4007-2213-4/2026/11}

\begin{document}

\title{DVAR: Adversarial Multi-Agent Debate \\for Video Authenticity Detection}


\author{Hongyuan Qi}
\email{qihy@zju.edu.cn}
\orcid{0009-0007-4096-6810}
\affiliation{%
  \department{College of Artificial Intelligence}
  \institution{Zhejiang University}
  \city{Hangzhou}
  \state{Zhejiang}
  \country{China}
}

\author{Feifei Shao}
\email{sff@zju.edu.cn}
\orcid{0000-0002-6384-9402}
\affiliation{%
  \department{State Key Lab of CAD\&CG}
  \institution{Zhejiang University}
  \city{Hangzhou}
  \state{Zhejiang}
  \country{China}
}

\author{Ming Li}
\correspondingauthor
\orcid{0000-0002-7852-0159}
\email{ming.li@u.nus.edu}
\affiliation{%
    \department{School of Artificial Intelligence}
  \institution{The Chinese University of Hong Kong, Shenzhen}
  \city{Shenzhen}
  \state{Guangdong}
  \country{China}
}

\author{Hehe Fan}
\email{hehe.fan.cs@gmail.com}
\orcid{0000-0001-9572-2345}
\affiliation{%
\department{State Key Lab of CAD\&CG}
  \institution{Zhejiang University}
  \city{Hangzhou}
  \state{Zhejiang}
  \country{China}
}

\author{Jun Xiao}
\email{Junx@zju.edu.cn}
\orcid{0000-0002-6142-9914}
\affiliation{%
\department{State Key Lab of CAD\&CG}
  \institution{Zhejiang University}
  \city{Hangzhou}
  \state{Zhejiang}
  \country{China}
}

\renewcommand{\shortauthors}{Qi et al.}
\begin{abstract}
The rapid evolution of video generation technologies poses a significant challenge to media forensics, as conventional detection methods often fail to generalize beyond their training distributions. To address this, we propose DVAR (Debate-based Video Authenticity Reasoning), a training-free framework that reformulates video detection as a structured multi-agent forensic reasoning process. Moving beyond the paradigm of pattern matching, DVAR orchestrates a competition between a Generative Hypothesis Agent and a Natural Mechanism Agent. Through iterative rounds of cross-examination, these agents defend their respective explanations against abnormal evidence, driving a logical convergence where the truth emerges from rigorous stress-testing. To adjudicate these conflicting claims, we apply Occam's Razor through the Minimum Description Length (MDL) framework, defining an Explanatory Cost to quantify the "logical burden" of each reasoning path. Furthermore, we integrate GenVideoKB, a dynamic knowledge repository that provides high-level reasoning heuristics on generative boundaries and failure modes. Extensive experiments demonstrate that DVAR achieves competitive performance against supervised state-of-the-art methods while exhibiting superior generalization to unseen generative architectures. By transforming detection into a transparent debate, DVAR provides explicit, interpretable reasoning traces for robust video authenticity assessment.

\end{abstract}



\maketitle
\begin{figure}[!t]
    \centering
    \includegraphics[width=0.9\columnwidth]{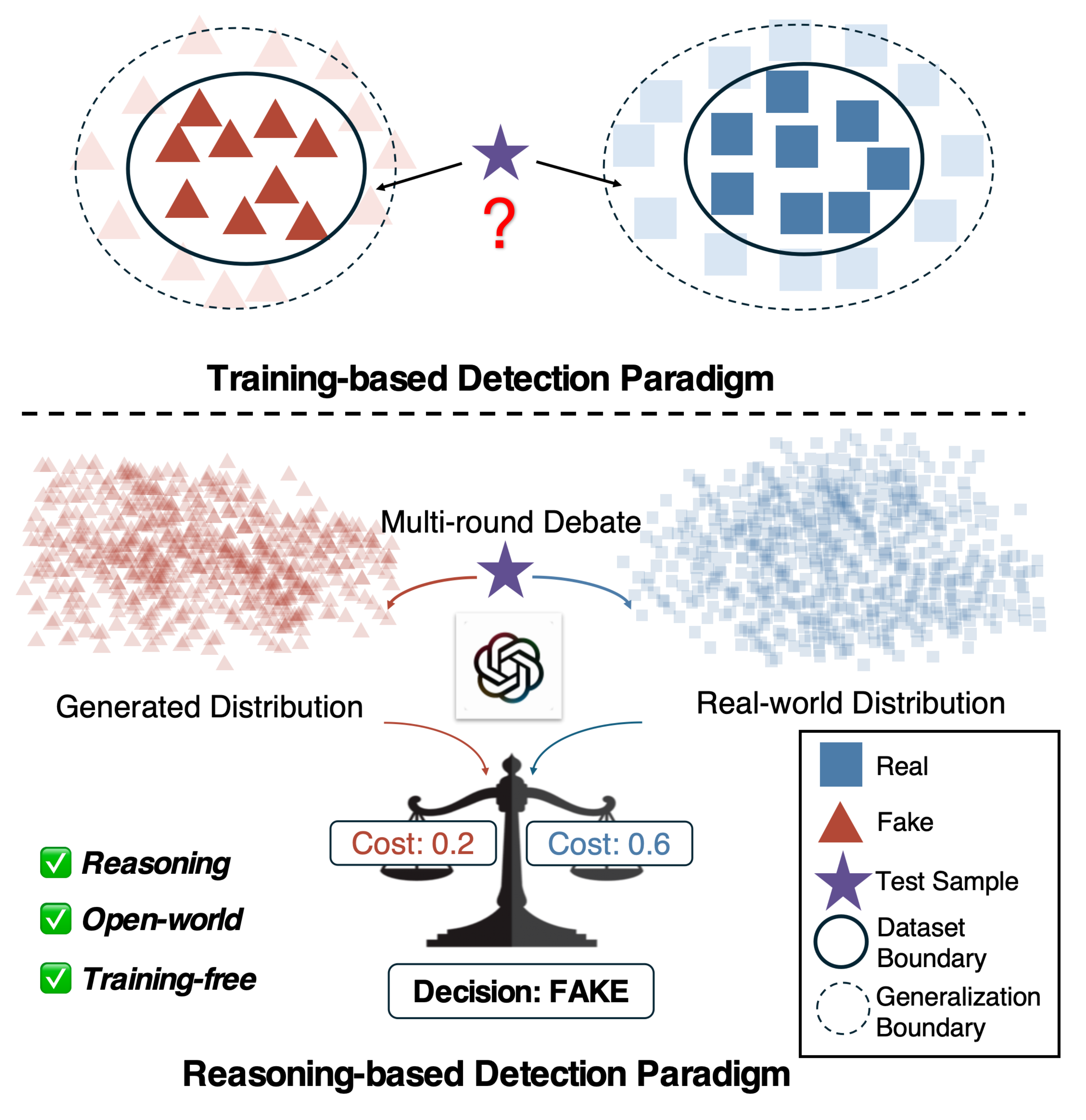}
   \caption{Training-based vs. Reasoning-based detection paradigms. 
Top: Training-based detection relies on pattern matching from curated datasets. Such black-box models are inherently bound to the training distribution, leading to a performance collapse when encountering unseen generative architectures.
Bottom: Reasoning-based detection (Ours) reformulates detection as a structured multi-agent debate. By quantifying the explanatory cost of competing hypotheses, DVAR enables training-free and interpretable authenticity assessment that generalizes to open-world scenarios.}
    \label{fig:teaser}
        \vspace{-6mm}
\end{figure}

\section{Introduction}
Recent advances in video generation models~\cite{liu2024sorareviewbackgroundtechnology,chen2024videocrafter2} have enabled the creation of highly realistic AI-generated videos across diverse scenarios with complex scenes, dynamic motion, and high visual fidelity, making synthetic videos increasingly difficult to distinguish from real content, fueling the spread of disinformation, privacy violations, and fraud~\cite{9712265, bateman2022deepfakes, westerlund2019emergence}. 
Consequently, reliable AI-generated video detection has become a critical problem~\cite{ho2020, dhariwal2021diffusionmodelsbeatgans}.

Existing deepfake detection methods primarily focus on learning artifact representations from labeled datasets~\cite{li2020facexraygeneralface, mccloskey2018detectinggangeneratedimageryusing, nataraj2019detectinggangeneratedfake, frank2020leveragingfrequencyanalysisdeep}. While effective on seen data, these black-box models often fail to generalize to unseen models in practice~\cite{ricker2024, song2023robustnessgeneralizabilitydeepfakedetection}. Moreover, due to the lack of explicit reasoning, they frequently rely on inscrutable correlations while overlooking blatant physical violations~\cite{ojha2023fakedetect,khan2024clipping,Tran2025Explainable,lin2025standingshouldersgiantsreprogramming}. Framing detection as a supervised pattern-matching task inherently limits robustness against rapidly evolving synthesis technologies, motivating a shift toward more transparent and reasoning-based forensics. Early studies have attempted to leverage Large Vision-Language Models (LVLMs) for forgery detection~\cite{zhang2024DDVQA,chakraborty2025truthlens}. However, even strong commercial models often underperform significantly, exhibiting suboptimal accuracy and unstable behavior~\cite{foteinopoulou2024hitchhikersguidefinegrainedface}.

Inspired by forensic practices, human investigators approach forgery detection from a fundamentally different perspective. 
Rather than relying on memorized artifact patterns, they examine suspicious content for physical anomalies that rarely occur in the natural world and reason about plausible explanations. 
For instance, an investigator may identify abnormal temporal motion or inconsistent lighting, subsequently evaluating whether these traces stem from natural processes or generative synthesis.
To bridge the gap between human intuition and machine perception, we propose a new detection paradigm that leverages LVLMs to simulate this cognitive process. This marks a paradigm shift from dataset-driven classification to a framework grounded in Evidence Discovery and Hypothesis Reasoning.

We argue that this limitation stems not from a lack of reasoning capability, but from a foundational misalignment in problem formulation. Specifically, LVLMs face three challenges adapt to detection tasks:\textbf{ (1) a granularity gap}, where models prioritize high-level semantics over subtle forensic cues;\textbf{ (2) reasoning hallucination}, leading to ungrounded explanations without structured constraints; and\textbf{ (3) hypothesis ambiguity}, where detected anomalies are incorrectly attributed to generative artifacts instead of natural physical processes. Consequently, effective video forensics requires more than raw reasoning power; it necessitates a structured mechanism to guide evidence discovery, enforce explanatory constraints, and explicitly adjudicate between competing hypotheses.

To address these challenges, we propose \textbf{DVAR} (Debate-based Video Authenticity Reasoning), a structured framework that reformulates video detection into a four-stage forensic reasoning process, anchored by an external knowledge repository.
Specifically, DVAR first isolates informative evidence from subtle physical anomalies that surpass high-level semantics. It then orchestrates a multi-agent adversarial debate to elicit competing explanations, effectively suppressing reasoning hallucinations. To adjudicate these claims, we introduce an MDL-based Explanatory Cost to evaluate hypotheses based on their logical economy, thereby distinguishing generative artifacts from natural physical processes. Finally, a global aggregation module synthesizes trace-level evidence into a consistent, unified verdict.
Crucially, DVAR integrates GenVideoKB, a dynamic knowledge base of generative mechanisms and failure modes, to provide steady-state heuristics for evidence interpretation. Together, these components shift deepfake detection from superficial pattern matching to a transparent, reasoning-driven paradigm, ensuring superior robustness and generalization in open-world scenarios.

Our contributions are summarized as follows:
\vspace{-1mm}
\begin{itemize}
    \item We propose DVAR, a multi-agent framework that reformulates video detection as a structured forensic reasoning problem. By adjudicating competing hypotheses through adversarial debate and MDL-based cost comparison, DVAR eliminates the reliance on task-specific datasets.

    \item We introduce GenVideoKB, a dynamic high-level reasoning heuristics. By incorporating generative boundaries and distilled failure modes, it mitigates reasoning hallucinations and enhances decision stability in open-world scenarios.

    \item Extensive experiments demonstrate that DVAR achieves competitive performance against state-of-the-art methods and exhibits superior generalization across unseen generative architectures without any task-specific training.
\end{itemize}

\section{Related Work}

\subsection{CNN-based AI-Generated Detection}
In the early stage of deep learning \cite{zu2025collaborative,cvpr/gait3d_v1,parsinggait_gps,li2026vlaattcadaptivetesttimecompute,li2026sentinelvlametacognitivevlamodel,He2023UnifiedSGGHOI,He2022OpenVocabularySGG,li2025multi,li2023ultrare,li2025survey,ACIL2022NeurIPS,Zhuang_2023_CVPR,AAAI2024DSAL},  research formulates deepfake detection as a representation learning problem, where visual artifacts are extracted and classified to distinguish authentic and manipulated content. CNN networks (\textit{e.g.}, Resnet~\cite{he2015deepresiduallearningimage}) typically serve as pattern extractors or classifiers.

FaceForensics++~\cite{roessler2019faceforensicspp} introduces a large-scale benchmark dataset and demonstrates the effectiveness of CNN-based detectors for detecting facial manipulations. Face X-ray~\cite{li2020facexraygeneralface} detects face forgeries by learning blending artifacts introduced during image compositing. 
DIRE~\cite{wang2023dire} computes reconstruction errors produced by diffusion inversion using DDIM and uses these discrepancies to identify diffusion-generated images. DeepShield~\cite{cai2025deepshield} aggregates local and global forgery cues within video frames to improve detection of manipulated videos. D3~\cite{yang2025d3} learns discrepancy representations between authentic and manipulated samples to enhance detection robustness.
These methods typically extract visual representations through feature transformation or neural networks and perform authenticity prediction using a classifier. Some approaches additionally provide localization of manipulated regions~\cite{li2020facexraygeneralface}.

Although these detectors achieve strong performance on known manipulation methods, their effectiveness is closely tied to the training distribution and may decrease when encountering generation models that are not covered by the training data. 
Moreover, the decision process of such detectors is typically difficult to interpret, which limits their reliability. 

\subsection{LVLM-based AI-Generated Detection}
Recent work explores LVLMs for AI-generated media detection by leveraging multimodal representations and cross-modal interactions. 
M2F2-Det~\cite{guo2025rethinking} integrates visual and textual representations to produce interpretable predictions for manipulated faces. 
CSCL~\cite{li2025unleashing} introduces a consistency learning strategy to align multimodal representations and improve detection robustness. 
CLIP-IFDL~\cite{NoiseAssisted} designs a noise-assisted prompt learning mechanism that guides vision-language models to focus on manipulation-sensitive cues.

Another line of research formulates deepfake detection as a visual question answering task. DDVQA~\cite{zhang2024DDVQA} constructs a large-scale question–answer dataset and trains multimodal models to answer authenticity-related questions about manipulated media. 
AIGI-Holmes~\cite{zhou2025aigi} employs multimodal large language models to analyze visual evidence and generate textual explanations for AI-generated image detection. 

These approaches leverage the multimodal representations and reasoning capabilities of large models to improve interpretability and semantic understanding. 
However, they still rely on dataset construction and model training or adaptation. Their generalizable performance remains closely tied to the available training data, which require continuous dataset expansion to accommodate emerging generation models.

In contrast to these approaches, our work formulates this task as a structured reasoning process and introduces a training-free framework designed for open-world AI-generated video detection.

\section{DVAR}
\label{sec:method}

\begin{figure*}[t]
    \centering
    \includegraphics[width=\textwidth]{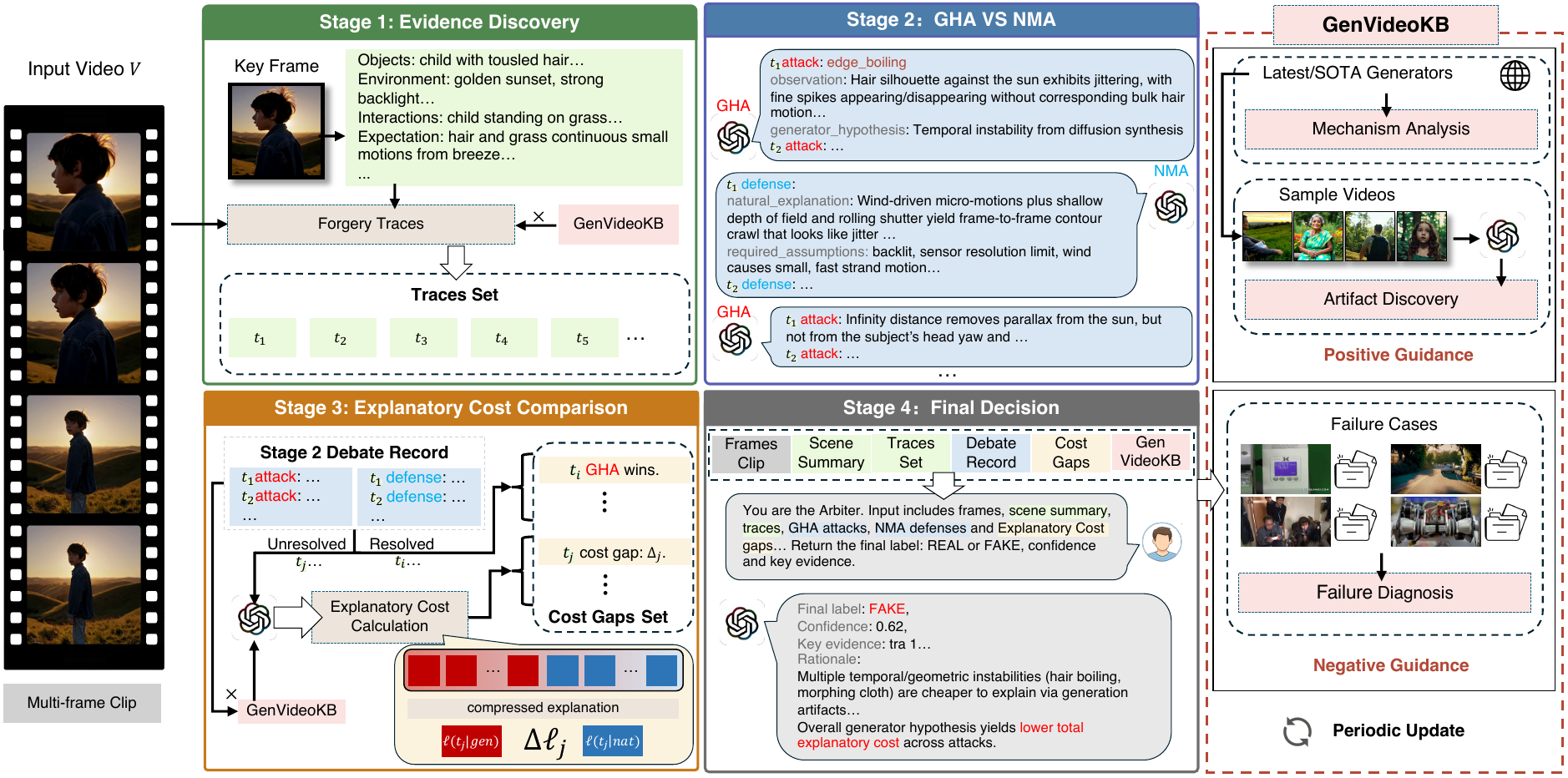}
    \caption{Overview of the DVAR framework. The pipeline consists of four stages: (1) \textbf{Evidence Discovery}, where semantic scenes and forgery traces are extracted; (2) \textbf{Hypothesis Debate}, where Generative Hypothesis Agent (GHA) and Natural Mechanism Agent (NMA) compete to explain traces; (3) \textbf{Explanatory Cost Comparison}, which quantifies the plausibility of each hypothesis based on assumption complexity; and (4) \textbf{Final Decision}, where an Arbiter aggregates reasoning signals. The \textit{GenVideoKB} supports all stages and evolves via failure diagnosis.}
    \label{fig:method}
\end{figure*}

Figure~\ref{fig:method} illustrates the architecture of the proposed DVAR framework. DVAR formulates video authenticity detection as a structured process consisting of four stages. 
Given an input video, the framework first identifies anomalies from sampled frames, and then constructs competing generative and natural explanations through a debate procedure. 
Subsequently, these explanations are evaluated through an explanatory cost calculation and a final decision is made by aggregating the resulting signals. 
GenVideoKB provides auxiliary knowledge in the form of semantic-level descriptions of generative capability boundaries and failure modes. 
We describe each component in detail below.

\subsection{Evidence Discovery}
\label{subsec:evidence}

Given an input video $\mathcal{V}$ with duration $S$, we uniformly sample frames at $n$ fps to construct a clip representation $\mathcal{C} = \{f_1, f_2, ..., f_T\}$, where $T = n \times S$. To capture semantic context, a key frame $f_{key} \in \mathcal{C}$ is selected for scene understanding.

A multimodal perception module analyzes $f_{key}$ and $\mathcal{C}$ to generate two outputs:
\begin{enumerate}
    \item \textbf{Scene Observation ($O_{scene}$)}: A structured description of the environment, objects, and interactions (\textit{e.g.}, \textit{``child standing on grass under golden sunset''}).
    \item \textbf{Forgery Traces ($\mathcal{T}$)}: A set of $M$ anomalies identified by comparing frame dynamics against physical expectations.
\end{enumerate}
Formally, the traces set is defined as $\mathcal{T} = \{t_1, t_2, ..., t_M\}$. Each trace $t_i$ represents an observable phenomenon requiring explanation. During this stage, GenVideoKB is used as auxiliary prior knowledge to support the identification and interpretation of temporally inconsistent or physically implausible phenomena.

\subsection{Multi-agent Multi-round Debate}
\label{subsec:debate}

Given the trace set $\mathcal{T}$, DVAR does not treat individual traces as direct indicators of forgery. 
Instead, it formulates explanation as a multi-round adversarial debate between competing hypotheses constructed by two reasoning agents: the \textit{Generative Hypothesis Agent (GHA)} and the \textit{Natural Mechanism Agent (NMA)}.

For each trace $t_i \in \mathcal{T}$, the agents construct competing explanations:
\begin{itemize}
    \item \textbf{GHA (Attack)}: Attributes $t_i$ to synthetic origins (\textit{e.g.}, diffusion instability, VAE compression). It proposes a generative hypothesis $H_{gen}^{(i)}$.
    \item \textbf{NMA (Defense)}: Attributes $t_i$ to natural physical processes (\textit{e.g.}, wind-induced motion, rolling shutter, depth of field). It proposes a natural hypothesis $H_{nat}^{(i)}$.
\end{itemize}

To sustain its stance, each agent must actively propose its own hypothesis, contest the opposing explanation with contradictory evidence, and rebut the opponent's challenges.This multi-agent adversarial interaction is specifically designed to mitigate individual hallucinations and compel each agent to logically converge toward a reliable, self-consistent explanation under its respective stance. The maximum depth of this adversarial process is controlled by a pre-defined hyperparameter of reasoning rounds. However, the debate may terminate early if one agent fails to effectively rebut a challenge or implicitly concedes when its hypothesis becomes insufficient to explain the observed phenomenon. This early termination avoids redundant reasoning steps and signifies that a clear explanatory preference has already emerged.Upon completion, the debate yields a structured record $D = \{d_1, \dots, d_M\}$, where each entry $d_i$ captures the competing hypotheses $(H_{gen}^{(i)}, H_{nat}^{(i)})$ alongside their respective auxiliary assumption sets $(\mathcal{A}_{gen}^{(i)}, \mathcal{A}_{nat}^{(i)})$. This ensures that only the most robust explanations survive to the final evaluation stage.

\subsection{Explanatory Calculation}
\label{subsec:cost}

Given the trace set $T = \{t_i\}_{i=1}^M$, DVAR adjudicates authenticity by comparing the logical economy of competing hypotheses. Guided by Occam’s Razor, we posit that the more plausible world model is the one requiring fewer auxiliary assumptions to explain the observed evidence. We operationalize this through an MDL-inspired framework, approximating explanatory complexity via normalized description length.

\textit{Decision Mechanisms} For each trace $t_i$, we first evaluate whether the adversarial debate yields a clear winner. A trace is resolved ($i \in \mathcal{R}$) if one agent fails to rebut a critical challenge or implicitly concedes, yielding a categorical signal:
\begin{equation}
r_i \in \{-1, +1\}, \quad i \in \mathcal{R}
\end{equation}
where $-1$ and $+1$ favor the natural and generative hypotheses, respectively. Otherwise, the trace is deemed unresolved ($j \in \mathcal{U}$) and subjected to compression-based evaluation.

\textit{Compression-based Explanation} To ensure stable cost estimation, we normalize unresolved hypotheses into a canonical form $E_m(t_j)$ via a constrained compression protocol:
\begin{equation}
E_m(t_j) = \text{Compress}(D_j, K_j, m), \quad m \in \{\text{nat}, \text{gen}\}
\end{equation}
where $D_j$ is debate trajectory and $K_j$ denotes forensic priors from GenVideoKB. To eliminate stochastic variance, we employ deterministic decoding ($temp=0$) under a fixed, structured template that mandates parsimony and penalizes redundancy. This ensures $E_m(t_j)$ preserves only essential causal factors.

\textit{Description-Length Cost} We define the explanatory cost $\ell(t_j \mid m)$ as the token length of the compressed explanation:
\begin{equation}
\ell(t_j \mid m) = \text{Length}(E_m(t_j)).
\end{equation}
The relative Cost Gap $\Delta \ell_j$ then quantifies the competitive advantage of one hypothesis over the other:
\begin{equation}
\Delta \ell_j = \ell(t_j \mid \text{nat}) - \ell(t_j \mid \text{gen}).
\end{equation}
By enforcing identical prompting and tokenization constraints, systematic biases cancel out, making the relative ordering of $\Delta \ell_j$ a robust proxy for explanatory efficiency.Evidence Aggregation. Finally, we aggregate signals from both resolved and unresolved traces into an evidence set $S$:
\begin{equation}
S = \{r_i\}_{i \in \mathcal{R}} \cup \{\Delta \ell_j\}_{j \in \mathcal{U}}.
\end{equation}
This structured collection serves as basis for global decision, ensuring the final verdict is grounded in granular, verifiable reasoning.
\subsection{Final Decision Arbiter}
\label{subsec:arbiter}
Given the set of trace-level signals, the final decision requires consolidating heterogeneous evidence into a global conclusion. 
Each trace contributes either a resolved decision $r_i$ or a cost-based signal $\Delta \ell_j$, and these signals may not always be fully consistent across traces. 
Therefore, a dedicated aggregation step is necessary to reconcile local evidence and produce a unified outcome.

We introduce an Arbiter module that performs constrained aggregation over the collected signals.
The Arbiter operates primarily on these structured trace-level signals, which encode the outcome of prior reasoning and cost evaluation. 
Its role is to consolidate them into a globally consistent decision and generate a structured output (\textit{e.g.}, label, confidence, and supporting evidence).
While auxiliary inputs such as frames and debate records are provided for interpretability, the final decision remains grounded in the structured signals.
To ensure stability and reproducibility, the Arbiter operates under a fixed prompt template with deterministic decoding, and its output is restricted to a predefined JSON schema. 

Overall, the Arbiter serves as a structured aggregation interface that consolidates trace-level evidence, rather than an independent decision-making component.
\subsection{GenVideoKB Construction and Evolution}
\label{subsec:kb}

GenVideoKB serves as an external knowledge base in DVAR, providing auxiliary semantic-level priors for trace-level adjudication. 

\paragraph{Knowledge representation.}
Each entry in GenVideoKB is represented as a structured tuple
\begin{equation}
k = (\text{phenomenon}, \text{description}, \text{type}),
\end{equation}
where \textit{phenomenon} denotes an observable visual or temporal effect (\textit{e.g.}, non-rigid deformation, temporal inconsistency), 
\textit{description} provides a natural language explanation, and 
\textit{type} indicates whether the entry provides \textbf{positive guidance} or \textbf{negative guidance}.
The KB is constructed at the level of observable phenomena rather than generator-specific profiles, ensuring that the knowledge remains generalizable across different systems.

\textit{Proactive knowledge acquisition (positive guidance).}
This stream focuses on summarizing the capabilities and limitations of modern generative models. 
We collect information from publicly available sources, including technical reports, official documentation, and representative outputs of state-of-the-art video generators. 
Descriptions of synthesis mechanisms and typical failure modes are distilled into semantic-level entries. 
These entries serve as \textit{positive guidance}, providing priors over plausible generative explanations grounded in known synthesis behaviors.

\textit{Reactive failure diagnosis (negative guidance).}
This stream refines the KB based on observed reasoning failures. 
Samples with incorrect predictions are analyzed to identify misleading cues or inconsistent reasoning patterns (\textit{e.g.}, over-attributing camera motion or lighting variations to generative processes). 
These cases are summarized as \textit{negative guidance}, which helps suppress unreliable interpretations and improves reasoning robustness over time.

\textit{Knowledge retrieval and usage.}
During inference, for each trace-level debate instance, we construct a query by concatenating trace description with debate context. 
This query is encoded into a dense representation using a pretrained text embedding model. 
Each KB entry is encoded in the same embedding space based on its phenomenon and description. 
We compute cosine similarity between the query and KB entries to retrieve the most relevant positive and negative entries.
The retrieved knowledge is provided as auxiliary context to adjudication module. 
Positive guidance supports explanations consistent with generative plausibility, while negative guidance discourages reasoning paths that have previously led to incorrect conclusions.

\textit{Update protocol and quality control.}
The KB is periodically updated through an automated pipeline. 
Candidate entries are filtered via consistency checks and deduplication, and a subset is manually verified to maintain quality. 
This hybrid strategy ensures scalability while preventing noise accumulation.

\textit{Discussion on prior and generalization.}
GenVideoKB introduces an explicit source of prior knowledge into DVAR. 
While the framework is training-free in that it does not rely on supervised learning over labeled datasets, it is not prior-free. 
These priors are expressed as abstract textual descriptions of observable phenomena, rather than dataset-specific visual patterns or generator identities. 
The KB does not store image-level features, latent representations, or benchmark annotations, and therefore does not encode dataset-specific information.
Consequently, GenVideoKB does not introduce information leakage from evaluation benchmarks. 
Instead, it provides a form of domain knowledge that supports consistent reasoning and improves the reliability of hypothesis comparison in open-world settings.

\section{Experiments}
\label{sec:experiments}

\begin{table*}[!t]
\vspace{-3mm}
\centering
\begin{minipage}{0.48\textwidth}
\centering \setlength{\tabcolsep}{17pt}
\caption{Performance comparison on Buster OOD benchmark. Supervised methods are trained on Buster train split.}
\label{tab:buster}

\begin{tabular}{|r||cc|}
\hline\thickhline
    \rowcolor{mygray}
Method & ACC (\%)  & F1 (\%) \\
\hline\hline
3D ResNet~\cite{hara2018spatiotemporal3dcnnsretrace} & 65.6 & 70.6 \\
ViViT~\cite{arnab2021vivitvideovisiontransformer} & 76.2 & 79.4 \\
DeMamba~\cite{DeMamba} & 79.3 & 82.0 \\
BusterX~\cite{wen2026busterxunifiedcrossmodalaigenerated} & 84.8 & 85.1 \\
BusterX++~\cite{wen2026busterxunifiedcrossmodalaigenerated} & \textbf{92.4} & 92.3 \\
\hline
GPT-5.4 (Direct) & 52.3 & 58.1 \\
GPT-5.4 (CoT) & 57.1 & 72.0 \\
\textbf{DVAR (Ours)} & 92.2 & \textbf{94.7} \\
\hline
\end{tabular}

\end{minipage}\hfill
\begin{minipage}{0.48\textwidth}
    \caption{Performance comparison on GenVidBench.}
    \label{tab:genvidbench}

    \centering \setlength{\tabcolsep}{6pt}
    \begin{tabular}{|r||cccc|}
\hline\thickhline
    \rowcolor{mygray}
&  \multicolumn{2}{c}{Full Benchmark} & \multicolumn{2}{c|}{Subset} \\
 \rowcolor{mygray}\multirow{-2}[-1]{*}{Method}& ACC (\%) & F1 (\%) & ACC (\%) & F1 (\%)  \\
\hline\hline
I3D~\cite{carreira2018quovadisactionrecognition} & 60.21 & 67.38 & 58.90 & 65.04 \\
SlowFast~\cite{feichtenhofer2019slowfastnetworksvideorecognition} & 70.06 & 81.13 & 72.44 & 82.73 \\
TIN~\cite{liu2021video} & 67.91 & 73.74 & 69.22 & 75.18 \\
TSM~\cite{lin2018temporal} & 73.88 & 82.49 & 75.69 & 83.48 \\
UniFormerV2~\cite{li2022uniformerv2} & 65.31 & 73.03 & 66.39 & 74.62 \\
VideoSwin~\cite{liu2021Swin} & 80.39 & 86.36 & 82.14 & 87.55 \\
MViTv2~\cite{li2021improved} & 80.45 & 85.66 & 79.32 & 84.74 \\
DeMamba~\cite{DeMamba} & 85.47 & 90.27 & 89.06 & 94.27 \\
\hline
\textbf{DVAR (Ours)} & -- & -- & \textbf{95.52} & \textbf{98.22} \\
\hline
\end{tabular}
\end{minipage}
\end{table*}
\subsection{Experimental Setup}
\label{subsec:setup}

\textbf{Datasets.}
We evaluate DVAR on two comprehensive benchmarks for AI-generated video detection: \textit{Buster}~\cite{wen2026busterxunifiedcrossmodalaigenerated} and \textit{GenVidBench}~\cite{ni2025genvidbenchchallengingbenchmarkdetecting}. 
\textbf{Buster} provides a long-form AI-generated video dataset and introduces an Out-of-Distribution (OOD) evaluation protocol that focuses on generalization to previously unseen generators. This setting aligns with DVAR's design goal of open-world detection without task-specific training. 
\textbf{GenVidBench} is a large-scale benchmark containing approximately 6.78 million videos generated by a diverse set of video generation models, covering a wide range of scenarios and motion complexities. 

\textbf{Baselines.}
We compare DVAR with representative deepfake detection models reported in the Buster and GenVidBench benchmarks. 
These include CNN-based architectures (\textit{e.g.}, 3D ResNet~\cite{hara2018spatiotemporal3dcnnsretrace}, I3D~\cite{carreira2018quovadisactionrecognition}), transformer-based models (\textit{e.g.}, ViViT~\cite{arnab2021vivitvideovisiontransformer}, TimeSformer~\cite{gberta_2021_ICML}, and recent state-of-the-art architectures such as DeMamba~\cite{DeMamba} and the BusterX~\cite{wen2026busterxunifiedcrossmodalaigenerated} series. 
Following the evaluation protocols of the respective benchmarks, the performance of supervised baselines is reported from their original publications~\cite{wen2026busterxunifiedcrossmodalaigenerated,ni2025genvidbenchchallengingbenchmarkdetecting} where available. 
In addition, we include two prompting baselines using the same multimodal backbone: \textit{Direct GPT} and \textit{Direct GPT+CoT} (chain-of-thought reasoning), in order to isolate the contribution of our structured reasoning framework beyond naive prompting.

\textbf{Implementation Details.}
Unless otherwise specified, all experiments use \textit{GPT-5.4} as LVLM model. 
Video frames are uniformly sampled at 5 frames per second (fps) to construct input clip $\mathcal{C}$. 
Each debate agent (GHA and NMA) performs two rounds of iterative argumentation before the explanatory cost calculation and final decision produced by Arbiter agent. The selection of these hyperparameters will be discussed in detail in Section~\ref{subsec:ablation}.
All inference is conducted through API calls on a standard workstation without requiring dedicated GPU clusters, highlighting the practical advantage of training-free detection.

\textbf{Knowledge Base Construction and Leakage Control}
To prevent potential data leakage, the GenVideoKB used in all experiments is constructed exclusively from external and publicly available sources, including official documentation of generative models, technical reports, and independent evaluations. 
No benchmark datasets (\textit{e.g.}, GenVidBench or Buster), nor any samples or annotations derived from them, are used in the KB construction process, either directly or indirectly.
Furthermore, the KB remains strictly frozen throughout all experiments, with no updates or modifications based on evaluation data. 
This design prevents any form of test-time adaptation or information leakage.

\textbf{Evaluation Metrics.}
Following prior work~\cite{li2020facexraygeneralface,ni2025genvidbenchchallengingbenchmarkdetecting} in media forensics, we report classification \textbf{Accuracy (ACC)} and \textbf{F1 Score}. 
ACC reflects overall classification performance, while F1 Score, as the harmonic mean of precision and recall, provides a balanced evaluation under potential class imbalance.
\subsection{Main Results}
\label{subsec:main_results}

\textbf{OOD Generalization on Buster.}
Table~\ref{tab:buster} reports the performance on the Buster OOD benchmark. 
All supervised detectors are trained on the Buster training split and evaluated on generators that are not observed during training. 
The results show that the performance of dataset-trained detectors varies substantially under this OOD setting. 
Early architectures such as 3D ResNet~\cite{hara2018spatiotemporal3dcnnsretrace} achieve only 65.6\% accuracy, while more recent models including ViViT~\cite{arnab2021vivitvideovisiontransformer} and DeMamba~\cite{DeMamba} reach 76.2\% and 79.3\%, respectively. 
The BusterX series further improves robustness through large-scale training and reinforcement learning, achieving 92.4\% accuracy with BusterX++.

Despite operating in a training-free setting, DVAR achieves 92.2\%/94.7\% scores, reaching performance comparable to BusterX++. 
This result highlights the effectiveness of DVAR.

\textbf{Large-Scale Performance on GenVidBench.}
To further validate the generality of our approach, we additionally evaluate DVAR on GenVidBench (Table~\ref{tab:genvidbench}). 
While Buster provides a carefully designed OOD protocol, relying on a single benchmark may still limit the evaluation scope. 
GenVidBench complements Buster by covering a broader set of video generators and generation pipelines, with diverse prompts, synthesis protocols, and scene compositions. 
Evaluating on this benchmark thus provides a complementary perspective on open-world generalization.
Due to the computational cost of API-based inference, we evaluate DVAR on a stratified subset of GenVidBench. 
The subset contains 4,126 videos, sampled to preserve the original class distribution and generator diversity, ensuring proportional coverage across different generation pipelines. 

To ensure that the subset provides a fair and representative evaluation, we analyze its consistency with full benchmark from multiple perspectives. 
First, the distribution of samples across different generators is preserved in the subset, ensuring that it reflects the same generative diversity as full benchmark. 
Second, the evaluation results remain consistent between the two settings: the relative ranking of baseline methods is largely preserved, and the performance differences are small without exhibiting systematic bias toward either higher or lower accuracy. 
These observations indicate that the sampled subset provides a representative approximation of the original benchmark in terms of both distribution and difficulty, enabling reliable evaluation under practical computational constraints.

Most existing detectors are trained directly on GenVidBench, whereas DVAR operates in a fully training-free manner without access to benchmark-specific training data. 
Despite this difference, DVAR achieves an accuracy of \textbf{95.52\%} on the sampled subset, outperforming the supervised DeMamba model evaluated under the same setting. 
These results suggest that reasoning-based detection can generalize beyond dataset-specific artifact patterns and remain effective under diverse generation conditions.

\textbf{Comparison with Prompting Baselines.}
We further compare DVAR with two prompting baselines using the same backbone model (GPT-5.4). 
Direct prompting performs poorly, achieving only 52.3\% accuracy. 
Introducing chain-of-thought (CoT) reasoning yields only limited improvement (57.1\% ACC), but still fails to reliably detect generated videos. The above experimental results are consistent with prior studies~\cite{foteinopoulou2024hitchhikersguidefinegrainedface}, indicating that directly applying LVLMs to deepfake detection is not feasible. 

In contrast, DVAR substantially improves performance under the same backbone by organizing the detection process into structured reasoning stages. 
This result suggests that the performance gain primarily arises from the proposed reasoning framework, rather than the underlying model capacity. 
Prompt templates for both baselines are provided in the supplementary material.

\subsection{Backbone Robustness}
\label{subsec:backbone}

To evaluate the generalizability of DVAR and ensure its effectiveness is not contingent on specific model iterations, we instantiated the framework using three LVLM backbones: \textit{GPT-4o}, \textit{GPT-5.4}, and \textit{Qwen-3.5-Plus}. All evaluations were conducted on Buster OOD benchmark with a fixed reasoning pipeline.

The results in Table~\ref{tab:backbone} demonstrate that DVAR maintains robust performance across different backbones, confirming its model-agnostic design. While more recent model iterations naturally yield marginal absolute improvements due to enhanced semantic grounding, the framework's relative effectiveness remains stable.

Notably, our investigation into failure cases uncovers the source of performance discrepancies. We found that the gap between backbones stems largely from their handling of the \textit{real} data distribution.  By design, DVAR initiates reasoning based on suspicious cues, which inherently establishes a \textit{forgery hypothesis prior}. When instantiated with backbones that possess less refined real-world knowledge, this prior can lead to \textit{over-sensitivity}, where natural variations in authentic content are misinterpreted as manipulation artifacts.

However, DVAR effectively mitigates this inherent sensitivity. Through mechanisms such as multi-round debate, knowledge base correction, and hypothesis cost evaluation, the framework recalibrates the decision boundary, preventing premature conclusions based on ambiguous cues. Consequently, DVAR instantiated on current backbones achieves performance competitive with state-of-the-art training-based detectors, validating that our reasoning-driven approach can match supervised methods without requiring task-specific fine-tuning.
\vspace{-2mm}
\begin{table}[htbp]
\centering
\caption{Backbone robustness of DVAR evaluated on Buster OOD benchmark.}
\label{tab:backbone}

\centering \setlength{\tabcolsep}{17pt}
\begin{tabular}{|r||cc|}
\hline\thickhline
    \rowcolor{mygray}
Backbone Model & ACC (\%) & F1 (\%)\\
\hline\hline
GPT-4o & 86.13 & 87.16 \\
GPT-5.4 & 92.20 & 94.70 \\
Qwen-3.5-Plus & 95.88 & 97.16 \\
\hline
\end{tabular}
\end{table}

\subsection{Ablation Study}
\label{subsec:ablation}

\textbf{Component Contribution.} To validate the design philosophy of DVAR and quantify the contribution of each reasoning module, we conducted a comprehensive ablation study on the Buster OOD benchmark. The results are summarized in Table~\ref{tab:ablation_component}.

\textit{1) Necessity of Structured Reasoning.} The baseline configuration, which relies on direct prompting without intermediate reasoning stages, yields near-random performance (52.30\% ACC). This starkly indicates that off-the-shelf VLMs lack the intrinsic capability to implicitly detect deepfake artifacts without explicit guidance.

\textit{2) Grounding and Logical Rigor.} Introducing \textbf{Evidence Discovery} stage yields a substantial gain (+16.83\% ACC), as it forces the model to ground its judgment on specific temporal traces rather than global impressions. Subsequently, \textbf{Hypothesis Debate} module further boosts performance (+10.85\% ACC) by mitigating confirmation bias; explicitly contrasting generative vs. natural explanations prevents DVAR from prematurely converging on a single hypothesis.

\textit{3) Quantitative Calibration and Knowledge Injection.} While the \textbf{Explanatory Cost Comparison} module offers a marginal gain in accuracy, it significantly improves the F1 score (+4.03\% over Debate-only), suggesting it enhances decision stability by quantitatively resolving ambiguous cues. Finally, integrating \textbf{GenVideoKB} further improves performance (+5.01\% ACC), indicating that general VLM knowledge alone is insufficient for forgery detection and that external priors on generative artifacts provide important complementary guidance.

Collectively, the progressive improvements demonstrate that DVAR's components are not merely additive but synergistic, transforming a weak baseline into a robust detector competitive with supervised SOTA methods.

\begin{table}[htbp]
\centering
\caption{Ablation study on the key reasoning components of DVAR. \textbf{Bold} indicates the best performance.}
\label{tab:ablation_component}
\centering \setlength{\tabcolsep}{6pt}
\resizebox{\columnwidth}{!}{
\begin{tabular}{|cccc||cc|}
\hline\thickhline
\rowcolor{mygray}Evidence & Debate & Cost & GenVideoKB & ACC (\%) & F1 (\%)\\
\hline\hline
\ding{55} & \ding{55} & \ding{55} & \ding{55} & 52.30 & 58.10 \\
\ding{51} & \ding{55} & \ding{55} & \ding{55} & 69.13 & 77.09 \\
\ding{51} & \ding{51} & \ding{55} & \ding{55} & 79.98 & 84.04 \\
\ding{51} & \ding{51} & \ding{51} & \ding{55} & 87.19 & 91.07 \\
\ding{51} & \ding{51} & \ding{51} & \ding{51} & \textbf{92.20} & \textbf{94.70} \\
\hline
\end{tabular}
}
\end{table}

\begin{figure*}[t]
    \centering
    \includegraphics[width=0.9\textwidth]{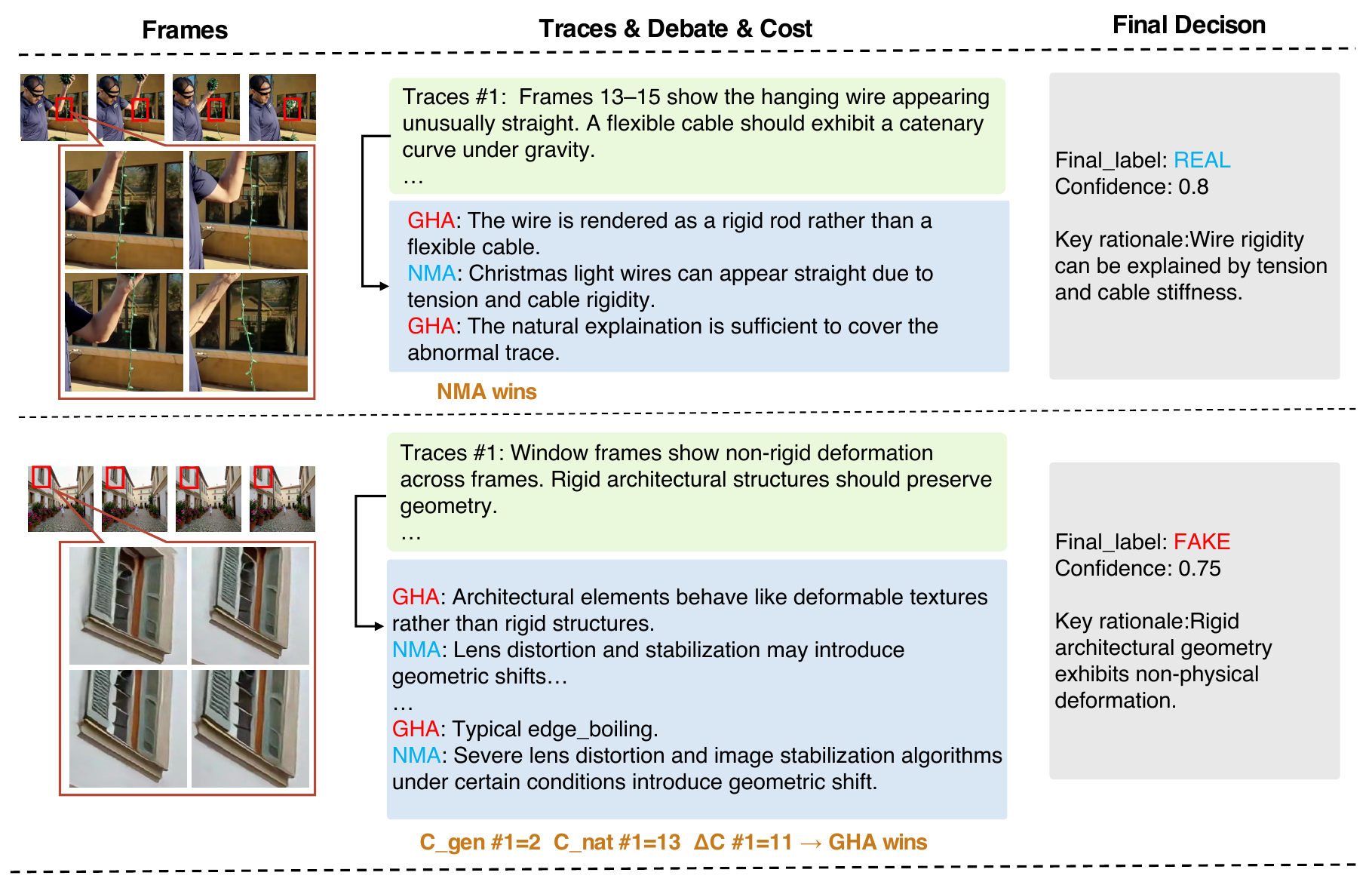}
    \caption{Case study illustrating the reasoning-driven detection process of DVAR. 
For each identified trace, the system adjudicates authenticity through a structured multi-agent debate. 
\textbf{Top (Real Case):} GHA acknowledges NMA's explanation for the detected anomaly before the debate concludes, leading to an early-terminated \textit{resolved} trace. 
\textbf{Bottom (Fake Case):} Both agents maintain their stances, but the explanatory cost for GHA is significantly lower than that of NMA, indicating a generative origin. 
Note that for visualization clarity, only selected rounds of a single representative trace are shown for each case; the actual decision is aggregated from multiple traces and comprehensive debate trajectories.}
    \label{fig:casestudy}
\end{figure*}
\textbf{Parameter Sensitivity.} The efficiency and robustness of DVAR are governed by two critical hyperparameters: the temporal granularity of input evidence (frame sampling rate) and the depth of logical reasoning (debate rounds). We conduct a sensitivity analysis to identify the optimal trade-off between performance gain and computational cost, as detailed in Table~\ref{tab:ablation_param}.

\textit{Temporal Granularity.} The sampling rate directly influences the model's ability to capture transient forgery artifacts. As shown in the top block of Table~\ref{tab:ablation_param}, increasing the rate from 1 fps to 5 fps yields a dramatic performance leap (+19.88\% ACC). This indicates that sparse sampling (1 fps) fails to preserve critical temporal dynamics essential for detection. However, further increasing to 8 fps provides only marginal gains (+0.91\% ACC) while incurring higher computational overhead, suggesting that 5 fps strikes the optimal balance between information density and redundancy.

\textit{Reasoning Depth.} The number of debate rounds determines the thoroughness of hypothesis verification. Comparing the bottom block with the baseline (5 fps, 2 rounds), we observe that expanding from 1 to 2 rounds significantly improves accuracy (+8.18\% ACC), confirming that a single pass is insufficient for rigorous logical critique. Performance saturates beyond 2 rounds (\textit{e.g.}, 3 rounds yield only +0.70\% ACC), indicating diminishing returns where additional reasoning steps introduce redundancy rather than new insights.

Based on these findings, we configure DVAR with \textbf{5 fps} and \textbf{2 debate rounds} as the default setting. This configuration maximizes detection capability while maintaining computational efficiency.

\begin{table}[htbp]
\centering
\caption{Sensitivity analysis of key reasoning parameters. The default configuration (5 fps, 2 rounds) is in gray.}
\label{tab:ablation_param}
\centering \setlength{\tabcolsep}{17pt}
\begin{tabular}{|cc||cc|}
\hline\thickhline
\rowcolor{mygray}
FPS & Rounds & ACC (\%) & F1 (\%) \\
\hline\hline
1 & 2 & 72.32 & 84.20 \\
\rowcolor{gray!15} 5 & 2 & \textbf{92.20} & \textbf{94.70} \\ 
8 & 2 & 93.11 & 94.31 \\
\hline
5 & 1 & 84.02 & 90.17 \\
\rowcolor{gray!15} 5 & 2 & \textbf{92.20} & \textbf{94.70} \\ 
5 & 3 & 92.90 & 94.59 \\
5 & 5 & 93.11 & 94.17 \\
\hline
\end{tabular}

\end{table}

\subsection{Computational Efficiency}
\label{subsec:efficiency}

While DVAR eliminates the need for costly training, it introduces additional inference overhead due to multi-stage reasoning and agent interactions. 
As shown in Table~\ref{tab:compute}, DVAR incurs approximately $3\times$ higher token consumption compared with direct prompting, primarily due to the debate and cost evaluation stages. 

However, a substantial portion of input tokens can be reused via caching (numbers in parentheses), which reduces the effective cost in practice. 
In particular, later stages benefit from high cache reuse, alleviating the overhead introduced by multi-round reasoning.

Despite the increased token usage, DVAR avoids expensive dataset collection and model training required by supervised detectors, making it attractive for rapidly evolving open-world scenarios.For practical deployment, a cascade strategy can be adopted, where lightweight detectors filter out trivial cases and only ambiguous samples are forwarded to DVAR for detailed reasoning. 
Reducing token consumption while maintaining performance remains an important direction for future work.

\begin{table}[t]
\centering
\small
\setlength{\tabcolsep}{5pt}
\renewcommand{\arraystretch}{1.15}
\caption{Average token consumption on Buster-OOD. Cached tokens are shown in parentheses.}
\label{tab:compute}
\centering \setlength{\tabcolsep}{6pt}
\begin{tabular}{|r|l||cc|c|}
\hline\thickhline
\rowcolor{mygray}
Method & Stage & Input & Output & Total \\
\hline\hline
Direct 
& -- 
& 16,210 {\scriptsize(0)} 
& 392 
& 16,602 {\scriptsize(0)} \\

CoT
& -- 
& 16,488 {\scriptsize(0)} 
& 1,377 
& 17,865 {\scriptsize(0)} \\

\hline

\multirow{4}{*}{DVAR}
& Stage 1 
& 16,490 {\scriptsize(0)} 
& 932 
& 17,422 {\scriptsize(0)} \\

& Stage 2 
& 16,990 {\scriptsize(16,103)} 
& 1,800 
& 18,790 {\scriptsize(16,103)} \\

& Stage 3 
& 18,814 {\scriptsize(16,256)} 
& 904 
& 19,718 {\scriptsize(16,256)} \\

& Stage 4 
& 19,579 {\scriptsize(18,442)} 
& 1,224 
& 20,803 {\scriptsize(18,442)} \\

\hline

\rowcolor{gray!20}
\textbf{DVAR (Total)} 
& -- 
& \textbf{76,533} {\scriptsize(50,801)} 
& \textbf{4,860} 
& \textbf{81,393} {\scriptsize(50,801)} \\

\hline
\end{tabular}
\end{table}








\subsection{Case Study.}
\label{subsec:qualitative}

We present representative examples to illustrate the reasoning process of DVAR.
As shown in Figure~\ref{fig:casestudy}, the system first identifies evidence such as \textit{unusually straight.} For readability, we present a condensed version of the reasoning process, focusing on the most representative traces and key hypothesis comparisons. 
The debate stage evaluates competing explanations: natural hypothesis requires environmental assumptions (\textit{e.g.}, wind and rolling shutter), whereas generative hypothesis attributes the phenomenon to diffusion instability. 
This reasoning trace provides interpretable evidence for the final decision.

\textbf{Failure Analysis.}
DVAR encounters difficulties in scenarios with limited temporal and physical variability. 
In particular, videos characterized by simple and repetitive motions (\textit{e.g.}, steady walking cycles), short durations, and minimal object interactions provide insufficient cues for reliable evidence discovery. 
In such cases, the lack of diverse temporal dynamics and cross-object constraints reduces the availability of discriminative traces, making it challenging for the reasoning process to distinguish between authentic recordings and high-quality generated content. 
This may lead to occasional false negatives. 
Additionally, highly compressed real videos may introduce artifacts that resemble generative inconsistencies, further complicating hypothesis evaluation. 
Future updates to GenVideoKB aim to address these limitations by incorporating priors tailored to low-motion scenarios and improving robustness to compression-induced distortions through failure diagnosis loop.
\section{Conclusion}
\label{sec:conclusion}
We present \textbf{DVAR}, a framework that reformulates video detection as interpretable structured reasoning. Through multi-agent debate and explanatory cost evaluation, DVAR provides transparent decision justifications while achieving robust performance. The integrated \textit{GenVideoKB} enables adaptation to unseen generators via proactive profiling and reactive diagnosis. We believe our work offers a sustainable path for trustworthy multimedia forensics against evolving generative AI.

\section*{Acknowledgement}
This work is supported by the National Natural Science Foundation of China (92570101, 62502317, 62441617), 
the Guangdong Basic and Applied Basic Research Foundation (Grant No. 2026A1515011198), 
Key R\&D Program of Zhejiang (2025C01128), 
Zhejiang Provincial Natural Science Foundation of China (No. LD25F020001), 
Fundamental Research Funds for the Central Universities (226-2025-00057), 
Postdoctoral Fellowship Program of CPSF (GZC20251077) and 
the Earth System Big Data Platform of the School of Earth Sciences, Zhejiang University.

\bibliographystyle{ACM-Reference-Format}
\bibliography{sample-base}
\end{document}